\documentclass[10pt,twocolumn,letterpaper]{article}

\usepackage{multirow}
\usepackage[pagenumbers]{cvpr} % To force page numbers, e.g. for an arXiv version

\definecolor{cvprblue}{rgb}{0.21,0.49,0.74}
\usepackage[pagebackref,breaklinks,colorlinks,allcolors=cvprblue]{hyperref}

\def\confName{CVPR}
\def\confYear{2026}

\title{Motion-Driven Multi-Object Tracking of Model Organisms in Space Science Experiments}

\author{
Jianing You$^{1,2}$ \quad 
Han Wang$^{1,2}$ \quad 
Kang Liu$^{1}$ \quad 
Jiale Ding$^{1,2}$ \\
Fengjie Chu$^{1,2}$ \quad 
Zihan Guo$^{1,2}$ \quad 
Shengyang Li$^{1,2}$\thanks{Corresponding author} \\
$^1$Technology and Engineering Center for Space Utilization, Chinese Academy of Sciences\\
$^2$School of Space Exploration, University of Chinese Academy of Sciences\\
{\tt\small \{youjianing24, liukang, wanghan22, shyli\}@csu.ac.cn}
}

\begin{document}
\maketitle
\begin{abstract}
Automated animal behavior analysis relies on long-term, interpretable individual trajectories; however, multi-animal tracking in space science experimental videos remains highly challenging due to weak appearance cues, low-quality imaging, complex maneuvering behaviors, and frequent interactions. To address this problem, we first construct the SpaceAnimal-MOT dataset to characterize the motion complexity and long-term identity preservation challenges in biological videos acquired under microgravity conditions. We then propose ART-Track (Adaptive Robust Tracking), a motion-driven tracking framework tailored to this setting. Specifically, multi-model motion estimation is introduced to handle abrupt maneuvers and nonlinear motion, motion-state-driven association is designed to reduce identity switches under dense interactions and temporary mismatch, and uncertainty-adaptive fusion is used to dynamically balance spatial and motion cues when prediction reliability varies. Experimental results show that ART-Track significantly reduces identity switches on zebrafish and fruitfly sequences, while maintaining more stable association under occlusion, deformation, and high-density interactions, thereby providing a more reliable tracking foundation for downstream quantitative behavior analysis. The code is publicly available at \url{https://github.com/yyy7777777/ART_TRACK}.
\end{abstract}    
\section{Introduction}
\label{sec:intro}
\begin{figure}[t]
	\centering
	\includegraphics[width=1.0\linewidth]{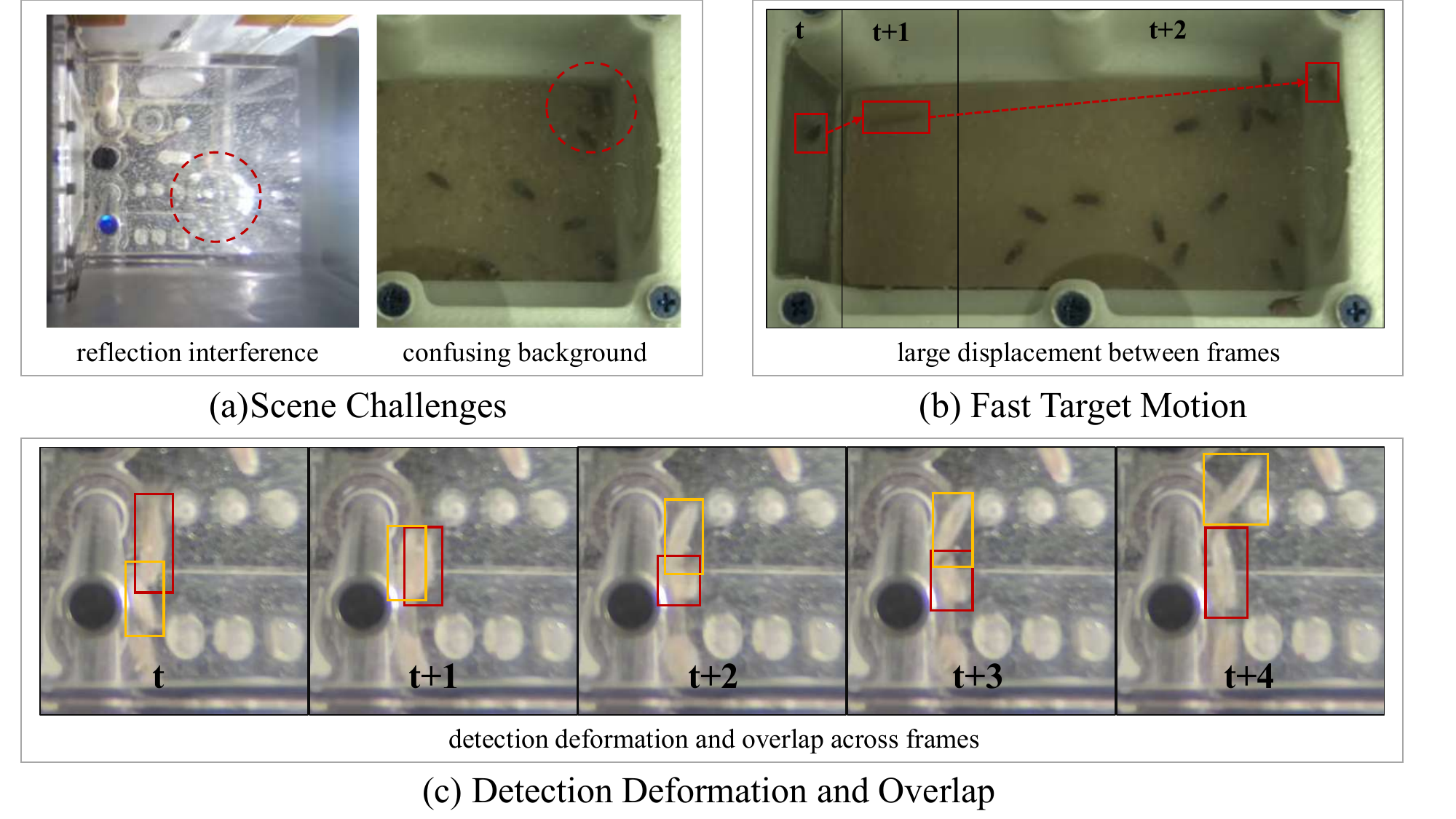}
	\caption{Typical challenges for tracking biological swarms in constrained experimental environments.}
	\label{datafeature}
\end{figure}
Model organisms such as zebrafish and fruitflies are widely used in life science, medicine, and space biology research. In microgravity experiments, long-term video observation of these animals helps reveal how environmental stress affects development, behavior, and physiology, while also providing an important reference for understanding potential impacts on astronaut health. In this context, multi-object tracking serves as a key step between raw observations and downstream behavior analysis. Because fully continuous trajectories without identity drift are often unrealistic, a more practical goal is to preserve usable trajectories for as long as possible while reducing identity switches, so as to support reliable quantification of behavioral phenotypes such as social interaction, activity level, and circadian rhythm.

However, tracking small model organisms in confined space-science experimental capsules remains challenging. As shown in Fig.~\ref{datafeature}, these videos present two major difficulties. On the one hand, small animals such as zebrafish and fruitflies move rapidly and flexibly, and microgravity further amplifies the nonlinearity of their motion, leading to abrupt displacement, sharp turns, and large deviations from simple linear dynamics. On the other hand, space-station observation conditions are constrained, so the recorded videos are often limited by low bitrate and low frame rate, together with reflections, bubble interference, and compression artifacts, making appearance cues weak and unreliable for association.

At the same time, the experimental setting has a closed-space characteristic: most targets remain within the observation chamber for long periods rather than frequently entering or leaving the scene. This shifts the main challenge from repeated target initialization to maintaining usable long-term trajectories and reducing identity switches after frequent interactions.

Our main contributions are summarized as follows:
\begin{itemize}
    \item We introduce SpaceAnimal-MOT, a multi-animal tracking dataset for microgravity videos, designed to highlight motion complexity, visual uncertainty, and long-term identity preservation in space science experimental settings.
    
    \item We propose ART-Track, a motion-driven tracking framework tailored for closed-space animal behavior videos, which is designed to maintain usable long-term trajectories and reduce identity switches under weak appearance cues, complex motion, and frequent interactions.
    
    \item Experimental results on zebrafish and fruitfly sequences show that ART-Track significantly reduces identity switches and achieves more stable association under occlusion, deformation, and high-density interactions, providing a more reliable basis for downstream quantitative behavior analysis.
\end{itemize}

\section{Dataset and Method Overview}
\label{sec:formatting}

%-------------------------------------------------------------------------
\subsection{Characteristics of SpaceAnimal-MOT Dataset}

\begin{figure}
    \centering
    
    \begin{subfigure}[b]{0.48\linewidth}
        \centering
        \includegraphics[width=\linewidth]{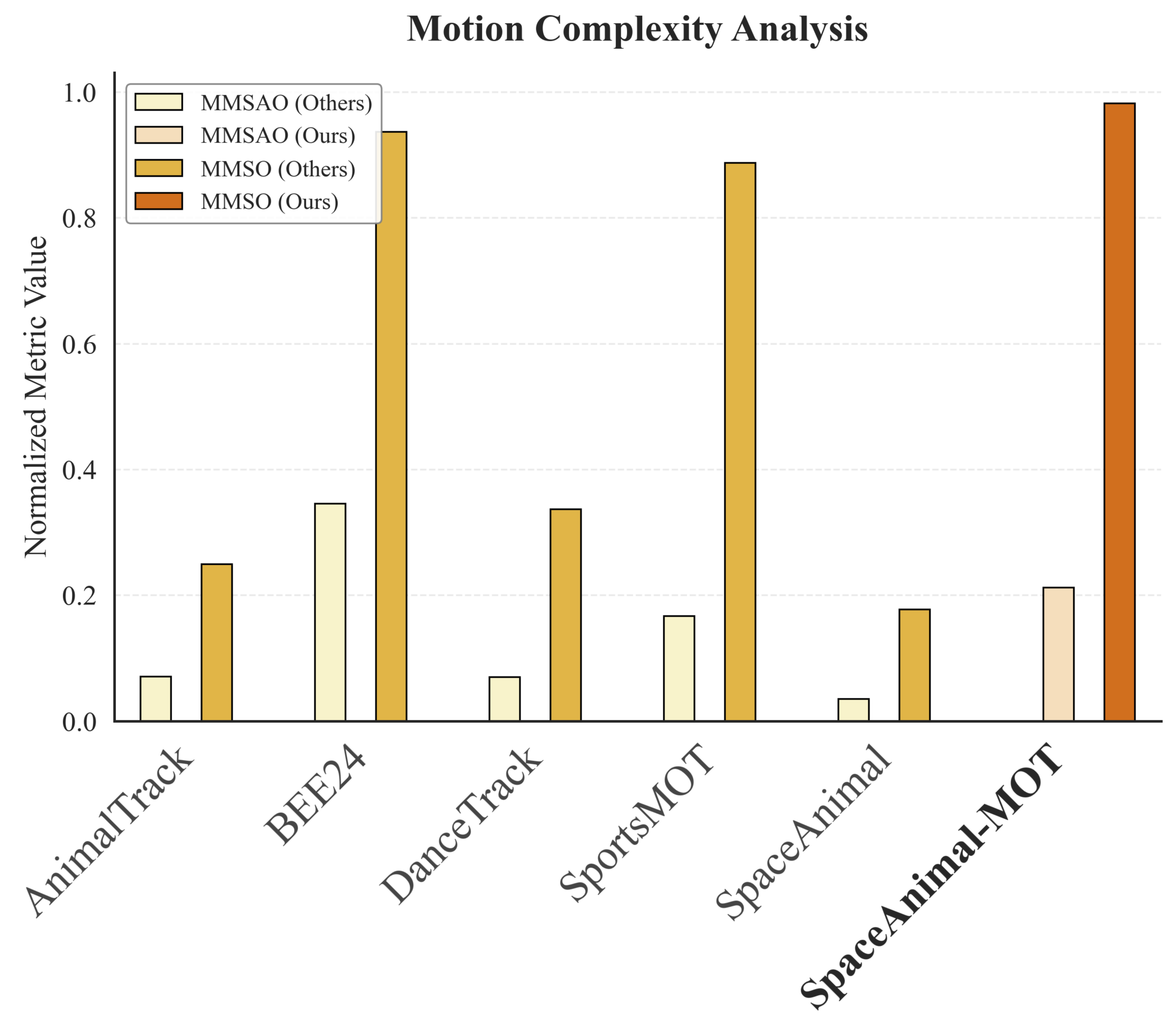}
        \caption{Motion Complexity Analysis}
        \label{fig:motion}
    \end{subfigure}
    \hfill
    \begin{subfigure}[b]{0.48\linewidth}
        \centering
        \includegraphics[width=\linewidth]{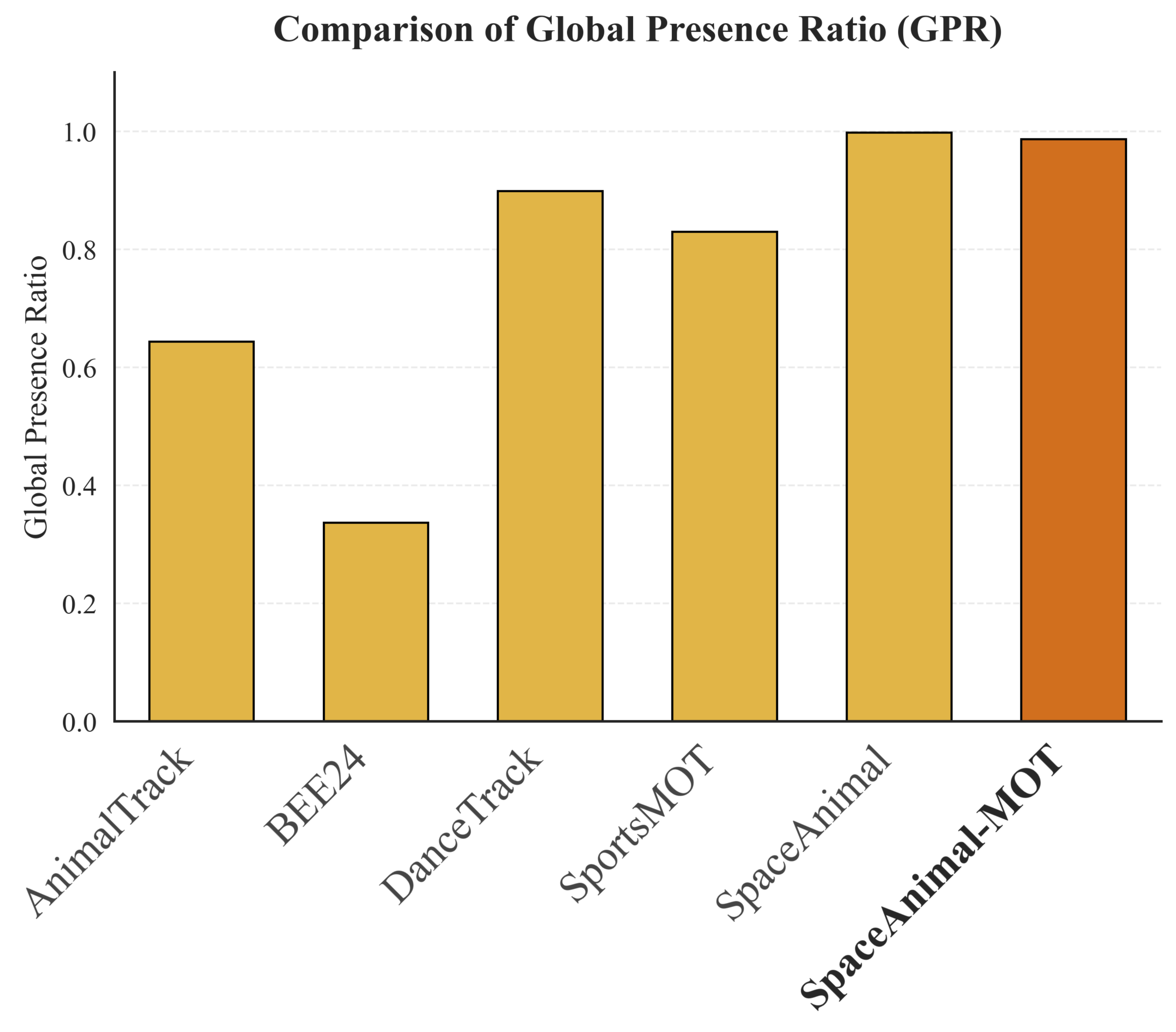}
        \caption{Global Presence Ratio (GPR)}
        \label{fig:gpr}
    \end{subfigure}
    
    \caption{Quantitative comparison analysis. (a) Motion complexity analysis (MMSAO and MMSO) highlights the dynamic challenges of our dataset compared to others. (b) The GPR metric demonstrates the high object presence frequency, indicating the closed-set nature of our dataset.}
    \label{fig:overall_comparison}
\end{figure}

To highlight the distinctive challenges of animal tracking in space science videos, we compare SpaceAnimal-MOT with mainstream tracking benchmarks, as shown in Fig.~\ref{fig:overall_comparison}. Unlike most benchmarks that mainly focus on occlusion or appearance similarity, SpaceAnimal-MOT places greater emphasis on complex motion and long-term identity preservation.

Specifically, the motion complexity analysis in Fig.~\ref{fig:motion} shows that SpaceAnimal-MOT exhibits substantially higher MMSO and MMSAO values than existing benchmarks~\cite{Cao2023TOPIC:}. This indicates that animals in our dataset frequently undergo abrupt speed changes, sharp turns, and nonlinear movements, while different individuals often display highly heterogeneous motion patterns at the same time. Therefore, reliable association in this setting depends less on appearance cues and more on motion modeling and temporal reasoning.

In addition, as illustrated in Fig.~\ref{fig:gpr}, SpaceAnimal-MOT has relatively long sequences and a Global Presence Ratio (GPR) close to 1.0, indicating that most targets remain visible for extended periods rather than frequently entering or leaving the scene. As a result, the main challenge of this dataset lies not in repeated target initialization, but in maintaining usable long-term trajectories and reducing identity switches after frequent interactions.

\begin{figure*}
    \centering
    \includegraphics[width=0.95\textwidth]{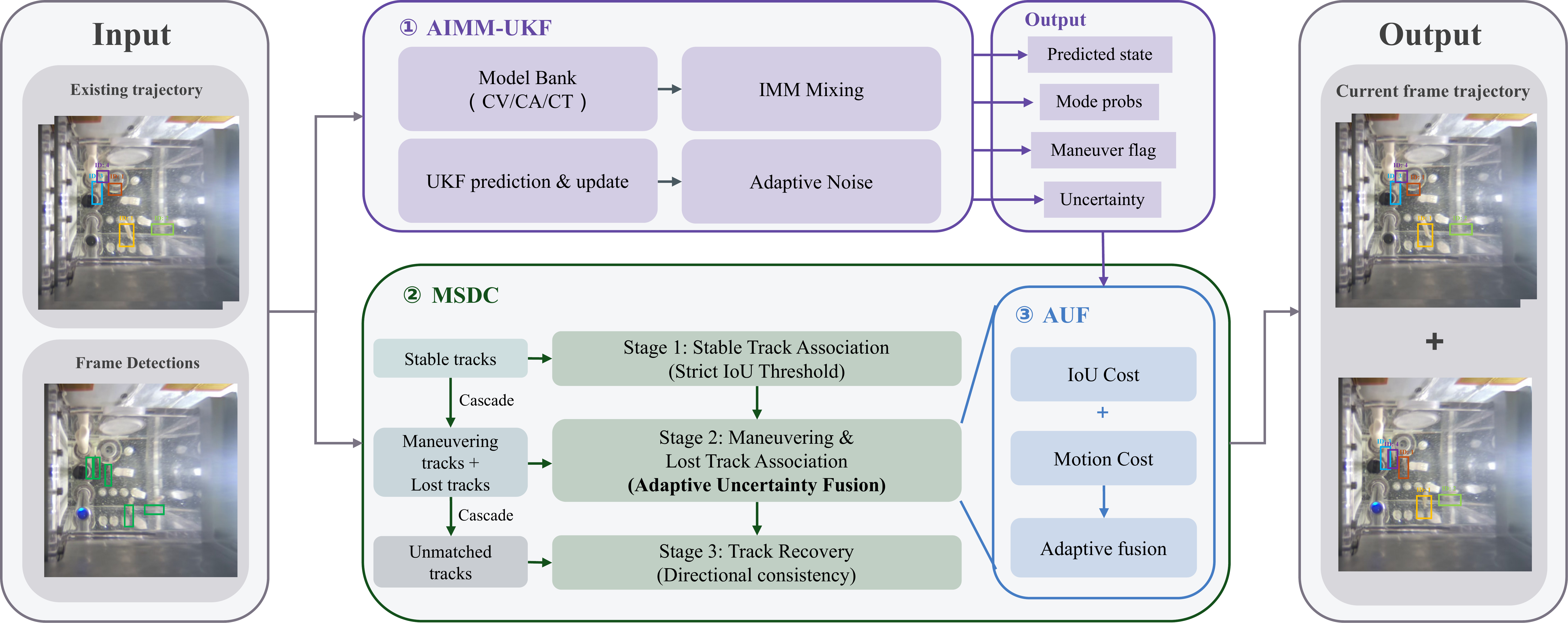}
    \caption{The overall architecture of the ART-Track framework.}
    \label{fig:framework}
\end{figure*}

\begin{table*}[ht]
\centering
\caption{Performance comparison on the \textbf{SpaceAnimal-MOT} dataset. Methods are grouped into \textit{Learnable Matchers}, \textit{Heuristic Matchers}, and \textit{Ours}. Since our method targets efficient heuristic-style association (learnable matchers are typically more computationally expensive), the \textbf{bold} and \underline{underlined} highlights denote the best and second-best results \textbf{only within the heuristic group (including ours)}. Learnable matchers are reported for reference without highlighting.}
\label{tab:comparison_with_learnable}

\scriptsize
\setlength{\tabcolsep}{2.5pt}
\renewcommand{\arraystretch}{1.08}

\resizebox{\textwidth}{!}{%
\begin{tabular}{l c c c c c c c c c c c c c}
\toprule
\multirow{2}{*}{\textbf{Tracker}} & \multirow{2}{*}{\textbf{Year}} &
\multicolumn{6}{c}{\textbf{SpaceAnimal-MOT (Zebrafish)}} &
\multicolumn{6}{c}{\textbf{SpaceAnimal-MOT (Fruitfly)}} \\
\cmidrule(lr){3-8} \cmidrule(lr){9-14}
& &
\textbf{HOTA}$\uparrow$ & \textbf{IDF1}$\uparrow$ & \textbf{IDs}$\downarrow$ & \textbf{MOTA}$\uparrow$ & \textbf{AssA}$\uparrow$ & \textbf{DetA}$\uparrow$ &
\textbf{HOTA}$\uparrow$ & \textbf{IDF1}$\uparrow$ & \textbf{IDs}$\downarrow$ & \textbf{MOTA}$\uparrow$ & \textbf{AssA}$\uparrow$ & \textbf{DetA}$\uparrow$ \\
\midrule

\multicolumn{14}{l}{\textit{Learnable Matchers}} \\
\hspace{1em}MOTR~\cite{zeng2021motr} & 2022 & 33.2 & 39.4 & 105 & 17.2 & 41.3 & 26.8 & 51.7 & 71.5 & 187 & 66.6 & 52.7 & 51.3 \\
\hspace{1em}MeMOTR~\cite{MeMOTR} & 2023 & 42.5 & 60.5 & 26 & 26.1 & 50.3 & 35.9 & 56.8 & 79.7 & 86 & 64.1 & 64.3 & 50.9 \\
\hspace{1em}SambaMOTR~\cite{segu2024samba} & 2024 & 41.9 & 58.3 & 27 & 20.6 & 51.2 & 34.3 & 56.2 & 77.5 & 91 & 60.3 & 64.4 & 50.7 \\
\hspace{1em}MOTIP~\cite{MOTIP} & 2025 & 52.0 & 66.3 & 855 & 74.4 & 43.5 & 62.2 & 54.6 & 72.7 & 2419 & 73.7 & 52.6 & 59.9 \\

\midrule
\multicolumn{14}{l}{\textit{Heuristic Matchers}} \\
\hspace{1em}Sort~\cite{Bewley2016} & 2016 & 15.0 & 20.1 & 246 & 23.2 & 11.8 & 19.1 & 52.1 & 70.0 & 295 & 67.2 & 53.6 & 51.8 \\
\hspace{1em}DeepSort~\cite{Wojke2017} & 2018 & 34.7 & 38.3 & 415 & 54.1 & 24.8 & 48.1 & 59.4 & 74.9 & 376 & 77.3 & 58.5 & 61.8 \\
\hspace{1em}FairMOT~\cite{Zhangetal2021} & 2021 & 35.8 & 43.4 & 472 & 53.2 & 27.1 & 47.8 & 50.6 & 64.9 & 849 & 63.5 & 50.6 & 53.7 \\
\hspace{1em}BOT-SORT~\cite{aharon2022bot} & 2021 & 39.5 & \underline{46.4} & 385 & \textbf{68.9} & 27.9 & \textbf{56.2} & 60.5 & \underline{79.3} & 491 & \underline{82.9} & \underline{61.7} & \underline{64.6} \\
\hspace{1em}CSTrack~\cite{cstrack} & 2022 & 18.6 & 21.6 & 451 & 27.9 & 12.4 & 28.1 & 49.2 & 63.5 & 880 & 64.2 & 49.9 & 52.5 \\
\hspace{1em}ByteTrack~\cite{Zhangetal2022} & 2022 & \underline{40.2} & 49.5 & \underline{171} & 57.4 & \underline{31.7} & \underline{51.4} & 61.1 & 78.5 & \underline{232} & 79.2 & 60.9 & 62.7 \\
\hspace{1em}OC-SORT~\cite{Cao2023} & 2023 & 38.6 & 46.8 & 180 & \underline{65.6} & 28.3 & 52.8 & \underline{61.8} & 78.9 & 227 & 82.8 & 61.0 & 63.7 \\
\hspace{1em}BoostTrack~\cite{stanojevic2024boostTrack} & 2024 & 37.2 & 45.5 & 342 & 58.7 & 28.9 & 48.1 & 61.1 & 76.9 & 328 & 80.6 & 60.5 & 62.9 \\
\hspace{1em}Hybrid-SORT~\cite{yang2024hybrid} & 2024 & 15.5 & 20.8 & 221 & 23.5 & 12.3 & 19.6 & 61.6 & 78.1 & 264 & 82.1 & 60.8 & 63.6 \\
\hspace{1em}FastTrack~\cite{hashempoor2025fasttrackerrealtimeaccuratevisual} & 2025 & 31.1 & 35.7 & 257 & 45.6 & 23.8 & 41.0 & 61.0 & 77.5 & 320 & 76.8 & 61.4 & 61.9 \\
\hspace{1em}SparseTrack~\cite{liu2025sparsetrack} & 2025 & 16.9 & 22.2 & 200 & 24.2 & 14.2 & 20.1 & 61.3 & \underline{79.3} & 237 & \underline{82.9} & \underline{61.7} & 64.2 \\

\midrule
\textbf{ART-Track (Ours)} & -- &
\textbf{40.5} & \textbf{56.8} & \textbf{26} & 61.3 & \textbf{33.4} & 49.2 &
\textbf{62.2} & \textbf{81.6} & \textbf{85} & \textbf{84.2} & 60.6 & \textbf{65.0} \\
\bottomrule
\end{tabular}%
}
\end{table*}

\begin{table*}[t!]
\centering
\caption{Upper-bound analysis on the \textbf{SpaceAnimal-MOT} dataset using \textbf{Oracle} (ground-truth) detections. The best results are highlighted in \textbf{bold}, and the second-best results are \underline{underlined}.}
\label{tab:oracle_comparison}

\scriptsize
\setlength{\tabcolsep}{2.5pt}
\renewcommand{\arraystretch}{1.08}

\resizebox{\textwidth}{!}{%
\begin{tabular}{l c c c c c c c c c c c c c}
\toprule
\multirow{2}{*}{\textbf{Tracker}} & \multirow{2}{*}{\textbf{Year}} &
\multicolumn{6}{c}{\textbf{SpaceAnimal-MOT (Zebrafish)}} &
\multicolumn{6}{c}{\textbf{SpaceAnimal-MOT (Fruitfly)}} \\
\cmidrule(lr){3-8} \cmidrule(lr){9-14}
& &
\textbf{HOTA}$\uparrow$ & \textbf{IDF1}$\uparrow$ & \textbf{IDs}$\downarrow$ & \textbf{MOTA}$\uparrow$ & \textbf{AssA}$\uparrow$ & \textbf{DetA}$\uparrow$ &
\textbf{HOTA}$\uparrow$ & \textbf{IDF1}$\uparrow$ & \textbf{IDs}$\downarrow$ & \textbf{MOTA}$\uparrow$ & \textbf{AssA}$\uparrow$ & \textbf{DetA}$\uparrow$ \\
\midrule

Sort~\cite{Bewley2016} & 2016
& 65.7 & 63.0 & 292 & 92.2 & 53.8 & 81.3
& 79.9 & 78.5 & 389 & 95.3 & 71.1 & 90.1 \\

DeepSort~\cite{Wojke2017} & 2018
& 53.9 & 50.0 & 427 & 81.2 & 40.2 & 73.8
& 75.7 & 75.2 & 419 & 92.9 & 66.9 & 86.1 \\

BOT-SORT~\cite{aharon2022bot} & 2021
& 79.5 & \underline{78.2} & 83 & \underline{99.2} & \underline{69.9} & 90.8
& 87.5 & 88.1 & 279 & \underline{99.2} & \underline{81.6} & 94.1 \\

ByteTrack~\cite{Zhangetal2022} & 2022
& 67.9 & 74.2 & \underline{73} & 84.2 & 62.9 & 74.4
& 83.1 & \underline{88.3} & 155 & 96.7 & 79.6 & 87.2 \\

OC-SORT~\cite{Cao2023} & 2023
& \underline{80.4} & 74.1 & 114 & 97.8 & 66.1 & \underline{98.0}
& \underline{88.9} & 85.5 & 228 & 98.0 & 80.4 & \underline{98.4} \\

BoostTrack~\cite{stanojevic2024boostTrack} & 2024
& 62.1 & 56.3 & 331 & 90.3 & 47.6 & 81.7
& 80.5 & 78.9 & 393 & 95.3 & 71.9 & 90.5 \\

Hybrid-SORT~\cite{yang2024hybrid} & 2024
& 72.4 & 63.3 & 212 & 95.6 & 54.4 & 96.3
& 84.1 & 80.0 & 324 & 95.8 & 73.3 & 96.4 \\

FastTrack~\cite{hashempoor2025fasttrackerrealtimeaccuratevisual} & 2025
& 67.2 & 72.8 & 79 & 83.0 & 62.4 & 73.5
& 81.9 & 86.6 & 179 & 95.9 & 77.9 & 86.5 \\

SparseTrack~\cite{liu2025sparsetrack} & 2025
& 77.5 & 76.4 & \underline{73} & 99.0 & 66.9 & 90.3
& 86.7 & 86.9 & \underline{159} & 99.0 & 80.4 & 93.6 \\

\midrule
\textbf{ART-Track (Ours)} & --
& \textbf{84.4} & \textbf{81.2} & \textbf{23} & \textbf{99.9} & \textbf{71.5} & \textbf{99.6}
& \textbf{93.6} & \textbf{93.7} & \textbf{83} & \textbf{99.7} & \textbf{87.9} & \textbf{99.8} \\

\bottomrule
\end{tabular}
}
\end{table*}

\begin{table}[t]
\centering
\caption{Ablation study of ART-Track on SpaceAnimal-MOT (Zebrafish).}
\label{tab:ablation_zebrafish}
\scriptsize
\setlength{\tabcolsep}{3pt}
\renewcommand{\arraystretch}{1.05}
\begin{tabular}{lccc cccc}
\toprule
\multirow{2}{*}{\textbf{Method}} & \multicolumn{3}{c}{\textbf{Components}} & \multicolumn{4}{c}{\textbf{Zebrafish}} \\
\cmidrule(lr){2-4} \cmidrule(lr){5-8}
 & AIMM-UKF & MSDC & AUF & HOTA$\uparrow$ & MOTA$\uparrow$ & IDF1$\uparrow$ & IDs$\downarrow$ \\
\midrule
Baseline   &  &  &  & 67.9 & 84.2 & 74.2 & 73 \\
           & \checkmark &  &  & 84.2 & 99.7 & 78.3 & 78 \\
           & \checkmark & \checkmark &  & 84.0 & 99.7 & 78.5 & 50 \\
           & \checkmark &  & \checkmark & 83.2 & 99.7 & 80.1 & 23 \\
\midrule
\textbf{ART-Track} & \checkmark & \checkmark & \checkmark & \textbf{84.4} & \textbf{99.9} & \textbf{81.2} & \textbf{23} \\
\bottomrule
\end{tabular}
\end{table}

\subsection{Overview of ART-Track Framework}
%-------------------------------------------------------------------------

ART-Track is a motion-driven tracking framework for complex animal behavior videos (Fig.~\ref{fig:framework}), designed to address the motion complexity and long-term association challenges revealed by the dataset statistics. Rather than relying on unreliable appearance cues in low-quality experimental videos, it follows a tracking-by-detection pipeline centered on motion-state estimation, dynamic association, and uncertainty modeling. It is particularly effective in closed-space experimental settings, where targets remain in view for long durations, allowing motion estimation to serve as a strong prior for recovering identities and maintaining long-term trajectories.

The first module, Adaptive Interacting Multiple Model Unscented Kalman Filter (AIMM-UKF), is designed to handle abrupt maneuvers and nonlinear motion in microgravity animal videos. Instead of assuming a single motion pattern, AIMM-UKF maintains three motion models in parallel, i.e., constant velocity (CV), constant acceleration (CA), and coordinated turn (CT), so that different motion regimes can be captured adaptively. The Unscented Kalman Filter propagates the state mean and covariance without explicit linearization, which makes it more suitable for sharp turns, sudden speed changes, and curvilinear motion. Given the mixed prediction from multiple models, AIMM-UKF outputs both the target state and its predictive uncertainty, which are later used to guide association.

Based on these motion estimates, the Motion State-Driven Cascaded (MSDC) Association Strategy addresses the problem of identity fragmentation under frequent interactions. Instead of matching all tracks with a single rule, MSDC first partitions trajectories into relatively stable and maneuvering tracks according to their motion states. Stable tracks are associated first with stricter IoU constraints, while maneuvering or temporarily lost tracks are handled in later stages with more tolerant matching conditions. This staged design reduces the chance that unstable trajectories dominate early assignment and therefore helps suppress identity switches in long sequences.

The third module, Adaptive Uncertainty Fusion (AUF), is introduced because the reliability of motion prediction changes over time. When the target undergoes abrupt maneuvers or temporary occlusion, fixed association weights can easily fail. AUF therefore dynamically fuses spatial and motion cues according to the uncertainty estimated by AIMM-UKF:
\[
C_{i,j}=\alpha_k C_{\mathrm{IoU}} + (1-\alpha_k) C_{\mathrm{mot}} \lambda_{\mathrm{state}},
\]
where $C_{\mathrm{IoU}}$ and $C_{\mathrm{mot}}$ denote the IoU-based and motion-based costs, respectively, $\alpha_k$ is an uncertainty-adaptive weight, and $\lambda_{\mathrm{state}}$ is a motion-state factor. When prediction uncertainty becomes large, the tracker relies more on spatial overlap; otherwise, stronger motion consistency is preserved.

Overall, ART-Track is designed not to pursue idealized error-free tracking, but to preserve usable trajectories for as long as possible and reduce identity switches in closed-scene space-science animal videos.

\section{Experiments}

\subsection{Experimental Setup and Core Results}
We evaluate ART-Track on the zebrafish and fruitfly subsets of SpaceAnimal-MOT. For fair comparison, all heuristic baselines and our method use detections from the same YOLOv11 detector. We report HOTA~\cite{Luiten2021HOTA}, IDF1~\cite{2016IDF1}, and the number of identity switches (IDs), where HOTA measures overall tracking quality, while IDF1 and IDs more directly reflect long-term identity preservation.

As shown in Table~\ref{tab:comparison_with_learnable}, ART-Track achieves more stable association on SpaceAnimal-MOT. On Zebrafish, it reaches an IDF1 of 56.8 and reduces IDs to 26, compared with 171 for ByteTrack and 180 for OC-SORT. On Fruitfly, it obtains the best HOTA (62.2) and IDF1 (81.6), while keeping IDs at 85. These results show that ART-Track more effectively reduces identity switches and improves long-term trajectory usability in complex animal videos.

\subsection{Upper-Bound Analysis and Ablation Studies}
To isolate the effect of detection errors, we further perform upper-bound analysis using ground-truth boxes. As shown in Table~\ref{tab:oracle_comparison}, under oracle detections, ART-Track achieves HOTA/IDF1 of 84.4/81.2 on Zebrafish and 93.6/93.7 on Fruitfly, with IDs of 23 and 83, respectively. This indicates that the proposed motion modeling and association design can maintain long-term trajectories more stably when detection noise is reduced.

We further study the contribution of each module through ablation experiments. As shown in Table~\ref{tab:ablation_zebrafish}, AIMM-UKF provides the largest overall gain, improving HOTA on Zebrafish from 67.9 to 84.2. MSDC then reduces IDs from 78 to 50 through motion-state-driven staged association, while AUF further lowers IDs to 23 by dynamically balancing spatial and motion cues according to predictive uncertainty. Overall, the performance of ART-Track comes from the complementary effects of multi-model state estimation, motion-state-driven association, and uncertainty-adaptive fusion. For brevity, we report the full ablation results on Zebrafish only, while the trends on Fruitfly are consistent.

\section{Conclusion}
\label{sec:rationale}
This paper studies multi-animal tracking in space-science experimental videos, where weak appearance cues, complex motion, and frequent interactions make association difficult. To address this problem, we introduce SpaceAnimal-MOT and develop ART-Track, a motion-centric framework for long-term tracking in animal experimental videos. Results on zebrafish and fruitfly sequences show that the proposed method reduces identity switches and improves long-term trajectory usability under occlusion, deformation, and dense interactions. While severe visual noise still leaves room for improvement, this work provides a stronger basis for downstream quantitative behavior analysis in space-science experiments. Future work will explore multi-camera settings to improve tracking accuracy and robustness.

{
    \small
    \bibliographystyle{ieeenat_fullname}
    \bibliography{main}

@INPROCEEDINGS{Bewley2016,
  author={Bewley, Alex and Ge, Zongyuan and Ott, Lionel and Ramos, Fabio and Upcroft, Ben},
  booktitle={2016 IEEE International Conference on Image Processing (ICIP)}, 
  title={Simple online and realtime tracking}, 
  year={2016},
  volume={},
  number={},
  pages={3464-3468},
  doi={10.1109/ICIP.2016.7533003}}

@INPROCEEDINGS{Wojke2017,
  author={Wojke, Nicolai and Bewley, Alex and Paulus, Dietrich},
  booktitle={2017 IEEE International Conference on Image Processing (ICIP)}, 
  title={Simple online and realtime tracking with a deep association metric}, 
  year={2017},
  volume={},
  number={},
  pages={3645-3649},
  doi={10.1109/ICIP.2017.8296962}}

@CONFERENCE{Zhangetal2022,
  author    = {Zhang, Y. and Sun, P. and Jiang, Y. and Yu, D. and Weng, F. and Yuan, Z. and Luo, P. and Liu, W. and Wang, X.},
  title     = {ByteTrack: Multi-object Tracking by Associating Every Detection Box},
  booktitle = {In Proc. Eur. Conf. Comput. Vis. (ECCV)},
  year      = {2022},
  pages     = {9-21}
}

@INPROCEEDINGS{Cao2023,
  author={Cao, Jinkun and Pang, Jiangmiao and Weng, Xinshuo and Khirodkar, Rawal and Kitani, Kris},
  booktitle={2023 IEEE/CVF Conference on Computer Vision and Pattern Recognition (CVPR)}, 
  title={Observation-Centric SORT: Rethinking SORT for Robust Multi-Object Tracking}, 
  year={2023},
  volume={},
  number={},
  pages={9686-9696},
  doi={10.1109/CVPR52729.2023.00934}}

@ARTICLE{Zhangetal2021,
  author  = {Zhang, Y. and Wang, C. and Wang, X. and Zeng, W. and Liu, W.},
  title   = {FairMOT: On the Fairness of Detection and Re-identification in Multiple Object Tracking},
  journal = {Int. J. Comput. Vis.},
  volume  = {129},
  year    = {2021},
  pages   = {3069-3087}
}

@article{aharon2022bot,
  title={BoT-SORT: Robust Associations Multi-Pedestrian Tracking},
  author={Aharon, Nir and Orfaig, Roy and Bobrovsky, Ben-Zion},
  journal={arXiv preprint arXiv:2206.14651},
  year={2022}
}

@article{Cao2023TOPIC:,
title={TOPIC: A Parallel Association Paradigm for Multi-Object Tracking Under Complex Motions and Diverse Scenes},
author={X. Cao and Y. Zheng and Yao Yao and Huapeng Qin and Xiaoyu Cao and Shihui Guo},
journal={IEEE Transactions on Image Processing},
year={2023},
volume={34},
pages={743-758},
doi={10.1109/tip.2025.3526066}
}

@InProceedings{MeMOTR,
    author    = {Gao, Ruopeng and Wang, Limin},
    title     = {{MeMOTR}: Long-Term Memory-Augmented Transformer for Multi-Object Tracking},
    booktitle = {Proceedings of the IEEE/CVF International Conference on Computer Vision (ICCV)},
    month     = {October},
    year      = {2023},
    pages     = {9901-9910}
}

@inproceedings{zeng2021motr,
  title={MOTR: End-to-End Multiple-Object Tracking with TRansformer},
  author={Zeng, Fangao and Dong, Bin and Zhang, Yuang and Wang, Tiancai and Zhang, Xiangyu and Wei, Yichen},
  booktitle={European Conference on Computer Vision (ECCV)},
  year={2022}
}

@article{segu2024samba,
  title={Samba: Synchronized Set-of-Sequences Modeling for Multiple Object Tracking},
  author={Segu, Mattia and Piccinelli, Luigi and Li, Siyuan and Yang, Yung-Hsu and Van Gool, Luc and Schiele, Bernt},
  journal={arXiv preprint arXiv:2410.01806},
  year={2024}
}

@InProceedings{MOTIP,
    author    = {Gao, Ruopeng and Qi, Ji and Wang, Limin},
    title     = {Multiple Object Tracking as ID Prediction},
    booktitle = {Proceedings of the Computer Vision and Pattern Recognition Conference (CVPR)},
    month     = {June},
    year      = {2025},
    pages     = {27883-27893}
}

@ARTICLE{cstrack,
  author={Liang, Chao and Zhang, Zhipeng and Zhou, Xue and Li, Bing and Zhu, Shuyuan and Hu, Weiming},
  journal={IEEE Transactions on Image Processing}, 
  title={Rethinking the Competition Between Detection and ReID in Multiobject Tracking}, 
  year={2022},
  volume={31},
  number={},
  pages={3182-3196},
  doi={10.1109/TIP.2022.3165376}}

@article{stanojevic2024boostTrack,
  title={BoostTrack: boosting the similarity measure and detection confidence for improved multiple object tracking},
  author={Stanojevic, Vukasin D and Todorovic, Branimir T},
  journal={Machine Vision and Applications},
  issn = {0932-8092},
  year={2024},
  volume={35},
  number = {3},
  doi={10.1007/s00138-024-01531-5}
}

@inproceedings{yang2024hybrid,
  title={Hybrid-sort: Weak cues matter for online multi-object tracking},
  author={Yang, Mingzhan and Han, Guangxin and Yan, Bin and Zhang, Wenhua and Qi, Jinqing and Lu, Huchuan and Wang, Dong},
  booktitle={Proceedings of the AAAI Conference on Artificial Intelligence},
  volume={38},
  number={7},
  pages={6504--6512},
  year={2024}
}

@misc{hashempoor2025fasttrackerrealtimeaccuratevisual,
      title={FastTracker: Real-Time and Accurate Visual Tracking}, 
      author={Hamidreza Hashempoor and Yu Dong Hwang},
      year={2025},
      eprint={2508.14370},
      archivePrefix={arXiv},
      primaryClass={cs.CV},
      url={https://arxiv.org/abs/2508.14370}, 
}

@article{liu2025sparsetrack,
  title={Sparsetrack: Multi-object tracking by performing scene decomposition based on pseudo-depth},
  author={Liu, Zelin and Wang, Xinggang and Wang, Cheng and Liu, Wenyu and Bai, Xiang},
  journal={IEEE Transactions on Circuits and Systems for Video Technology},
  year={2025},
  publisher={IEEE}
}

@article{luiten2021hota,
  title={HOTA: A Higher Order Metric for Evaluating Multi-Object Tracking},
  author={Luiten, Jonathon and Osep, Aljosa and Dendorfer, Patrick and Torr, Philip and Geiger, Andreas and Leal-Taix{\'e}, Laura and Leibe, Bastian},
  journal={International Journal of Computer Vision},
  volume={129},
  number={2},
  pages={548--578},
  year={2021},
  doi={10.1007/s11263-020-01375-2}
}

@article{2016IDF1,
  title={Performance Measures and a Data Set for Multi-Target, Multi-Camera Tracking},
  author={ Ristani, Ergys  and  Solera, Francesco  and  Zou, Roger S  and  Cucchiara, Rita  and  Tomasi, Carlo },
  journal={Springer, Cham},
  year={2016},
}
}


\end{document}